\documentclass[10pt,twocolumn,letterpaper]{article}

\usepackage{cvpr}
\usepackage{times}
\usepackage{epsfig}
\usepackage{graphicx}
\usepackage{amsmath}
\usepackage{amssymb}
\usepackage{floatrow}
\usepackage[breaklinks=true,bookmarks=false]{hyperref}

\cvprfinalcopy % *** Uncomment this line for the final submission

\def\cvprPaperID{****} % *** Enter the CVPR Paper ID here

\begin{document}

%%%%%%%%% TITLE
\title{Democratisation of Usable Machine Learning in Computer Vision}

\author{
Raymond Bond$^1$, Ansgar Koene$^4$, Alan Dix$^3$, Jennifer Boger$^5$, Maurice D. Mulvenna$^1$\\
 Mykola Galushka$^2$,Bethany Waterhouse Bradley$^1$, Fiona Browne$^1$, Hui Wang$^1$\\
Alexander Wong$^{5,6}$\\
$^1$Ulster University, Northern Ireland, UK\\
$^2$Auromind, Northern Ireland, UK\\
$^3$Swansea University, Wales, UK\\
$^4$University of Nottingham, Nottingham, UK\\
$^5$University of Waterloo, Ontario, Canada\\
$^6$DarwinAI, Ontario, Canada\\
}

\maketitle
%\thispagestyle{empty}

%%%%%%%%% ABSTRACT
\begin{abstract}
Many industries are now investing heavily in data science and automation to replace manual tasks and/or to help with decision making, especially in the realm of leveraging computer vision to automate many monitoring, inspection, and surveillance tasks. This has resulted in the emergence of the ‘data scientist’ who is conversant in statistical thinking, machine learning (ML), computer vision, and computer programming. However, as ML becomes more accessible to the general public and more aspects of ML become automated, applications leveraging computer vision are increasingly being created by non-experts with less opportunity for regulatory oversight. This points to the overall need for more educated responsibility for these lay-users of usable ML tools in order to mitigate potentially unethical ramifications.  In this paper, we undertake a SWOT analysis to study the strengths, weaknesses, opportunities, and threats of building usable ML tools for mass adoption for important areas leveraging ML such as computer vision. The paper proposes a set of data science literacy criteria for educating and supporting lay-users in the responsible development and deployment of ML applications.
\end{abstract}
\vspace{-0.12in}
%%%%%%%%% BODY TEXT
\section{Introduction}
\vspace{-0.12in}
The prevalence of digital technology is due to initiatives that sought to make them more accessible; for example, graphical interfaces substantially democratized the use of computers. Today, users of various abilities can employ end-user development (EUD) or end-user programming (EUP) tools to build their own computer programs using graphical programming environments, such as Simulink \cite{Dabney2004}, LabView \cite{Travis2007}, and Scratch \cite{Resnick2009}. Applications such as Clementine, WEKA \cite{Witten1999}, RapidMiner \cite{Hofmann2013}, and recently Ludwig~\cite{Ludwig2018} realised the idea of ‘interactive machine learning’; namely, the development of machine learning (ML) models without the user having to write computer code. However, there are several aspects of EUDs that limit their accessibility and use, particularly in the realm of computer vision. EUDs use technical nomenclature and the user experience (UX) of their interfaces have arguably not been optimised for lay-users. EUD tools normally run on desktop machines whereas the general public are increasingly becoming more comfortable with web-based tools for email, social media and office applications. Finally, these EUD tools do not automatically deal with data cleansing, wrangling, missing data imputation and therefore require the user to have some deeper understanding of the ML process.

The term ‘usable ML’ has been discussed by a small number of researchers, including \cite{Bailis2017} and \cite{Sarkar2015}; examples of this emerging field include human-centered ML and computer vision \cite{Sacha2016,Binnig2018}. More recent advances involve running ML in the cloud, which has been described as 'ML as a Service' (MLaaS or Cloud ML). A new initiative referred to as automated machine learning (AutoML) is evolving (e.g. Google AutoML, Auto-WEKA), where a user provides a dataset (in the case of AutoML, via a web-based interface) and a set of algorithms automatically performs a task that is normally completed by a data scientist, such as feature engineering, model selection, and optimisation. AutoML algorithms tests a large number of feature sets, hyperparameters and permutations of ML techniques allowing for the automatic creation of the 'best' model.  For example, a number of recent papers~\cite{nas,enas,mnas,gensynth} have demonstrated the ability to automatically build state-of-the-art deep neural networks for computer vision tasks such as image classification and object detection based on just image data.

While these initiatives point towards the increasing democratisation of accessible ML, particularly for computer vision, we propose that there is a usable ML spectrum (Figure 1). The decreasing reliance on experts as ML becomes more accessible begs for further exploration to distinguish between ‘usable’ as in doing and ‘usable’ as in understanding. Whilst making ML more accessible can be a force for good, it is important to consider potential negative ramifications. For example, the increased accessibility of usable ML tools will likely cause an increase in inadvertent unethical use of ML for computer vision because of ignorance of ML literacy amongst lay-users. Usable ML for computer vision is analogous to allowing people without knowledge of car mechanics to drive cars, and whilst this is the case, drivers do need to know how to drive a car and are expected to follow the rules of the road, and understanding the risks and hazards of driving. Likewise, usable ML should be complemented by a degree of data science literacy, particularly in the ethical use of ML bearing in mind the risks and hazards of ML deployment, especially in computer vision scenarios where the livelihood, safety, and well-being of individuals in society are at stake in applications ranging from security surveillance and manufacturing quality inspection to medical diagnosis and autonomous vehicles.

%\blindtext
\begin{figure}
%\begin{floatrow}
\begin{center}
\hspace{-0.2in}\ffigbox{%
  \includegraphics[width=8.7cm]{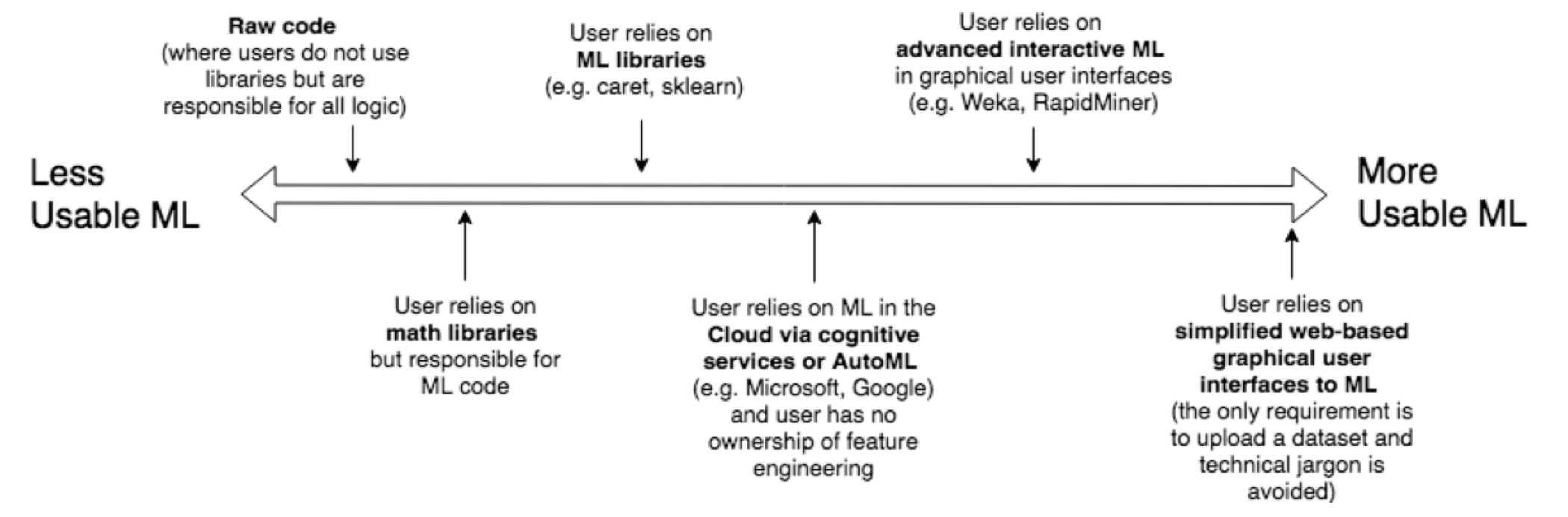}
}{%
  \caption{Spectrum of usable machine learning, from writing raw code (less usable) to web-based user interfaces (more usable).}%
}
~\\
\hspace{-0.2in}\ffigbox{%
  \includegraphics[width=8.7cm]{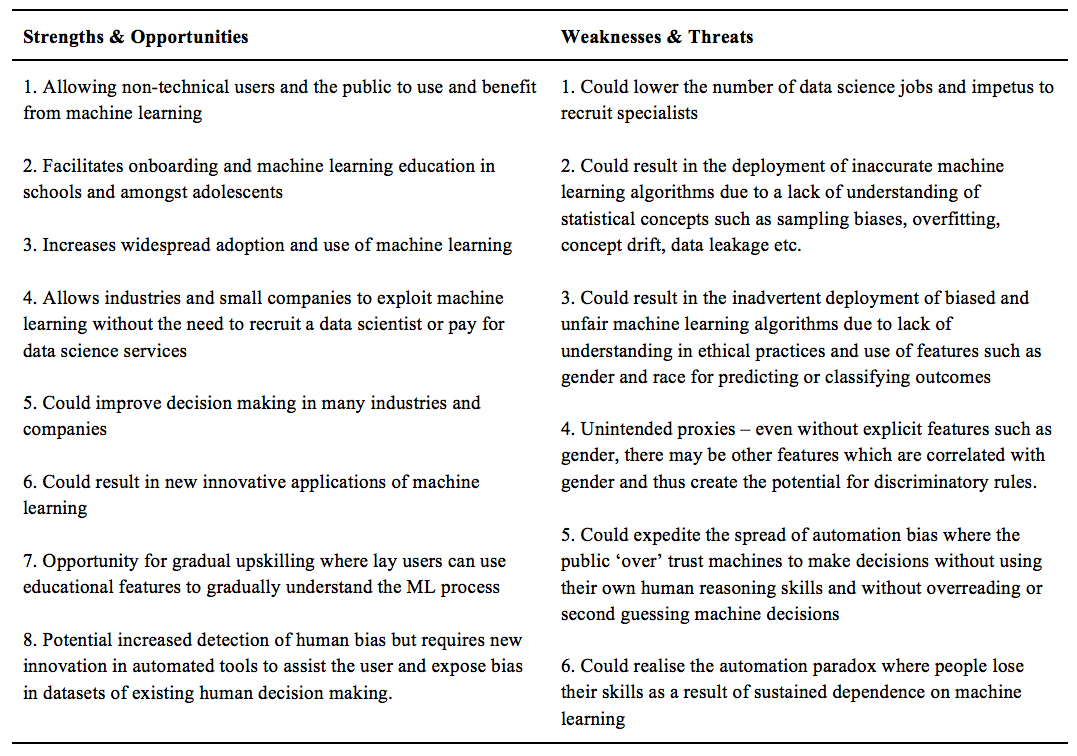}
}{%
  \caption{SWOT analysis of democratising usable ML}%
}
\end{center}
\vspace{-0.3in}
%\end{floatrow}
\end{figure}

If we want to support ethical algorithm development in computer vision, we need to ensure that this is possible, which requires the creation of elements missing from today’s usable ML tools. For example, an 'educational' feature with the potential for just-in-time learning of more theoretical ML concepts could suggest techniques or problems in the dataset whilst providing educational resources to provide the user the opportunity to have a greater level of understanding and ML literacy.

\section{SWOT Analysis}
\vspace{-0.12in}
We present a brief SWOT analysis of democratising usable ML for computer vision before outlining a working set of concepts that we believe are required to support  ethical and responsible use of ML platforms by lay-users (Table 1).

{\bf Strengths and Opportunities}:
Strengths and opportunities of democratising usable ML are presented in the left column of Table 1. This includes empowering the public and industries to use and deploy ML techniques for computer vision without the expense of a data scientist. It would certainly result in new innovative applications of ML in industries that were previously disadvantaged by a lack of technological investment and resource (for example, the agriculture industry is one that sees great benefit but is currently very behind in leveraging ML and computer vision). As a result, such usable ML tools could improve the decision making processes of many companies and verticals. Other benefits may include the widespread adoption of ML where it could be ubiquitous akin to spreadsheets. This would allow usable ML applications to be disseminated as part of primary and secondary education, which could result in a generation of 'ML-natives' who can leverage computer vision as a progression from digital natives (millennials).

{\bf Weaknesses and Threats }:
Weaknesses and threats of democratising usable ML are presented in the right column of Table 1. Democratisation could result in a reduced demand for data scientists. More worryingly, the democratisation of usable ML, especially in computer vision, could result in the inadvertent deployment of unethical ML algorithms that could perpetuate gender bias or racial discrimination unless they are audited and carefully developed by a qualified data scientist or statistician whilst using a set of ethical guidelines such as those being developed by the IEEE P7003TM standard working group (http://sites.ieee.org/sagroups-7003/). Naive users could release models that are very inaccurate for a number of reasons (e.g. sampling bias, etc.) that then result in models that poorly reflect the real-world. Naive users are generally also more prone to ‘automation bias’, where users put an imprudent amount of trust in ML algorithms that they believe to be accurate and become complacent, or do not think to second guess the algorithm’s predictions. In this sense, usable ML could cyclically accelerate automation bias. This and other considerations listed in Table 1 are just some of the possible risks of democratisating usable ML tools.

\section{Proposed benchmark criteria to form the basis of literacy in usable ML}
\vspace{-0.12in}
Considering the SWOT analysis in Table 1, we present a working set of benchmark criteria that could serve as data science literacy or certification for novice users of usable ML tools with the intention of mitigating irresponsible deployment of ML algorithms in computer vision.

{\bf Supervised ML} involves training an algorithm to learn patterns from a large dataset to be subsequently used to predict or classify an outcome when given new unseen cases (e.g., training a deep convolutional neural network to recognize faces with a large set of face images). While there are many techniques that can be used, the “no free lunch theorem” informs us that no one ML technique can be optimal for all problems and all domains. It is important that lay-users know that there are a multitude of ML techniques and algorithms and that care should be taken to select one that is appropriate for their data and context. This is important to avoid users overtrusting in one technique for all problems (e.g., deep learning) and to try different approaches to determine their relative strengths and weaknesses.

{\bf Accountability} refers to who is responsible for ML deployment and use. It is important that users understand that while an algorithm may be able to make autonomous decisions, the developers are becoming increasingly responsible for outcomes related to their use. Understanding this will enforce the seriousness of deploying ML algorithms.

{\bf Transparency} refers to how explainable and transparent an ML prediction is. Users must understand that different techniques provide different levels of transparency and explainability in understanding the inner workings or reasons that a particular prediction was made. This is important to allow users to use the right technique for the right domain and for their needs.  It is also important that users are aware that there are new tools and techniques available to provide improved explainability for techniques such as deep learning, which is traditionally viewed as a 'black box', for computer vision applications~\cite{clear}.

{\bf Data provenance} are the details, metadata and origin of the dataset used to build the ML model. It is important that the user knows that ML models are only as good as the data used to train them and that clear boundaries regarding appropriate use given the data used to train them should be considered and communicated.

{\bf Algorithmic bias} is when an ML model discriminates in some way (e.g., race, gender, age, etc.) \cite{Hajian2016}. It is important for the person developing the algorithm to identify, test for, and mitigate possible biases to ensure their algorithm should perform equitably across different populations; their algorithm should be fair.  New techniques are now available to assist in identifying these biases, and users should be aware of such tools to help to mitigate algorithmic bias.

{\bf Measurement bias} is when features or an outcome is poorly measured due to inexperience. Measurement bias could involve users only uploading data and features that are easily codified or inaccurately codified features, which results in suboptimal binning and categorization and thus ultimately in poor algorithm performance.

{\bf Accuracy vs. fairness:} An algorithm may produce results that are accurate but not fair by ethical standards. For example, there is a higher percentage of males in science and engineering, so in the absence of explicit information on qualifications - gender could be represented as a predictor of success in an engineering job; however, this would be unethical, unfair, and illegal (in the UK) to do so.  Without explicitly knowing and searching for these kinds of misrepresentations, a user may interpret accuracy figures reflect the best model to use.  This is particularly critical as ML for computer vision is increasingly leveraged by law enforcement for applications such as face recognition.

{\bf Automation bias} is when people place too much trust in decisions made by machines, sometimes to the point where they are complacent even when the machine is radically incorrect \cite{Bond2018}. It is important that those who deploy ML continuously question the decisions being made and keep in mind the accuracy of the model is not always correct.

{\bf Class imbalance and prevalence:} Prevalence is the percentage of cases that exist in the real world. Class imbalance is when there is an unwanted low percentage of a type of case (class) in the training dataset. It is important that users understand the limitations if the prevalence of cases in their dataset does not match up with the prevalence in the real world. In this way they can appreciate differentials in accuracies that will be achieved in the usable ML platform vs. the results achieved in real-world deployment.

{\bf Overfitting} is when a ML model performs well on the training data but not in the real world due to a ML model being so aligned to the training data that it is less generalizable when presented with new situations. It is important that users understand that while their model may have great performance during development, there is always a risk that it will not perform well in the real world due to overfitting.

{\bf Concept drift} is the phenomena that ML models may not always sustain the same performance over time, since predictors and circumstances can change over time. This concept is important to lay-users since they should understand the need to retrain models using more recent cases for certain disciplines (e.g., in autonomous vehicle applications; new vehicle types and models and new street sign types get introduced over time and have different visual appearance). Other concepts such as Goodhart's law (Chrystal et al. 2003) can explain concept drifts (i.e., once a variable is used as a measure to predict another variable, it can be manipulated insomuch that it is no longer a covariate).

{\bf Data leakage} is when predictors that are apparently random (e.g., house number, black bars used for image padding, etc.) are used in the ML model but seem to have predictive power. Data leakage occurs when the ‘answer’ is leaked inadvertently into training the ML algorithm and the performance is very high largely on test data due to the leaked feature but is very low in the real world. This concept is important to ensure users carefully consider what information should be included when training an ML algorithm.

{\bf A confounder} is a variable, feature, or predictor correlated to the variable the user wants to predict but may not be a lasting correlation due to it being a latent secondary or tertiary association. It is important that a user understands that confounders may allow ML to get the right answer for the wrong reason and that correlation is not causality.

{\bf ML performance metrics} include accuracy, sensitivity, specificity, mean average precision (mAP), and many others (e.g., kappa, area under the curve etc.). The accuracy paradox is when the ML algorithm achieves a misleadingly high accuracy score but is no better than prevalence rate of the most popular class in the dataset (also known as the 'no-information rate').  It is important for the user to understand accuracy and the accuracy paradox to guard against misleading themselves, clients, and colleagues.

{\bf Type 1 and type 2 errors:} Type 1 error is a false positive and a type 2 error is a false negative. This is important for users to understand so that they can have a better understanding of the ramifications of using an ML based decision and design for the error type that should be avoided the most. This will also help them choose an algorithm based on sensitivity or specificity depending on what the preference is in a given that domain or problem.
\vspace{-0.2in}
\section{Discussion and Conclusion}
\vspace{-0.12in}
The key message here is that "with great power comes great responsibility". Allowing development and deployment of ML applications, particularly for computer vision, to be more expedient, convenient, accessible, and more user friendly must be paired with methods that provide new and naive users with the knowledge that they need to be responsible actors. While not everyone wants to or can become experts in ML, responsible deployment still requires some literacy in ML and ethical concepts related to ML use.

While there are numerous positive outcomes that can come with democratising usable ML platforms for computer vision, we must guard against possible negative ramifications of widespread access to ML capabilities. We believe ML literacy that enables basic responsible use and deployment of ML models ought to be a paramount priority of usable ML systems. The concepts presented in this paper complement perspectives to other ethical positions, such as \cite{Mulvenna2017} and the IEEE P7003TM working standard, and provide a starting point for engaging in the responsible democratisation of usable ML.

{\small
\bibliographystyle{ieee}
\bibliography{egbib}

\begin{thebibliography}{10}\itemsep=-1pt

\bibitem{Bailis2017}
P.~Bailis, K.~Olukotun, C.~Ré, and M.~Zaharia.
\newblock Infrastructure for usable machine learning: The stanford dawn
  project.
\newblock 2017.

\bibitem{Binnig2018}
C.~Binnig, B.~Buratti, Y.~Chung, C.~Cousins, T.~Kraska, Z.~Shang, E.~Upfal,
  R.~C. Zeleznik, and E.~Zgraggen.
\newblock Towards interactive curation \& automatic tuning of ml pipelines.
\newblock In {\em DEEM@ SIGMOD}, pages 1--1, 2018.

\bibitem{Bond2018}
R.~R. Bond, T.~Novotny, I.~Andrsova, L.~Koc, M.~Sisakova, D.~Finlay,
  D.~Guldenring, J.~McLaughlin, A.~Peace, V.~McGilligan, et~al.
\newblock Automation bias in medicine: The influence of automated diagnoses on
  interpreter accuracy and uncertainty when reading electrocardiograms.
\newblock {\em Journal of electrocardiology}, 2018.

\bibitem{Dabney2004}
J.~B. Dabney and T.~L. Harman.
\newblock {\em Mastering simulink}.
\newblock Pearson, 2004.

\bibitem{Hajian2016}
S.~Hajian, F.~Bonchi, and C.~Castillo.
\newblock Algorithmic bias: From discrimination discovery to fairness-aware
  data mining.
\newblock In {\em Proceedings of the 22nd ACM SIGKDD international conference
  on knowledge discovery and data mining}, pages 2125--2126. ACM, 2016.

\bibitem{Hofmann2013}
M.~Hofmann and R.~Klinkenberg.
\newblock {\em RapidMiner: Data mining use cases and business analytics
  applications}.
\newblock 2013.

\bibitem{clear}
D.~Kumar, A.~Wong, and G.~W. Taylor.
\newblock Explaining the unexplained: A class-enhanced attentive response
  (clear) approach to understanding deep neural networks.
\newblock 2017.

\bibitem{Ludwig2018}
P.~Molino, Y.~Dudin, and S.~S. Miryala.
\newblock Introducing ludwig, a code-free deep learning toolbox.
\newblock 2017.

\bibitem{Mulvenna2017}
M.~Mulvenna, J.~Boger, and R.~Bond.
\newblock Ethical by design: A manifesto.
\newblock In {\em Proceedings of the European Conference on Cognitive
  Ergonomics 2017}, pages 51--54. ACM, 2017.

\bibitem{enas}
H.~Pham, M.~Y. Guan, B.~Zoph, Q.~V. Le, and J.~Dean.
\newblock Efficient neural architecture search via parameter sharing.
\newblock 2018.

\bibitem{Resnick2009}
M.~Resnick, J.~Maloney, A.~Monroy-Hern{\'a}ndez, N.~Rusk, E.~Eastmond,
  K.~Brennan, A.~Millner, E.~Rosenbaum, J.~Silver, B.~Silverman, et~al.
\newblock Scratch: programming for all.
\newblock {\em Communications of the ACM}, 52(11):60--67, 2009.

\bibitem{Sacha2016}
D.~Sacha, M.~Sedlmair, L.~Zhang, J.~A. Lee, D.~Weiskopf, S.~North, and D.~Keim.
\newblock Human-centered machine learning through interactive visualization.
\newblock ESANN, 2016.

\bibitem{Sarkar2015}
A.~Sarkar.
\newblock Spreadsheet interfaces for usable machine learning.
\newblock In {\em 2015 IEEE VL/HCC}, pages 283--284. IEEE, 2015.

\bibitem{mnas}
M.~Tan, B.~Chen, R.~Pang, V.~Vasudevan, and Q.~V. Le.
\newblock Mnasnet: Platform-aware neural architecture search for mobile.
\newblock 2018.

\bibitem{Travis2007}
J.~Travis and J.~Kring.
\newblock {\em LabVIEW for everyone: graphical programming made easy and fun}.
\newblock Prentice-Hall, 2007.

\bibitem{Witten1999}
I.~H. Witten, E.~Frank, L.~E. Trigg, M.~A. Hall, G.~Holmes, and S.~J.
  Cunningham.
\newblock Weka: Practical machine learning tools and techniques with java
  implementations.
\newblock 1999.

\bibitem{gensynth}
A.~Wong, M.~J. Shafiee, B.~Chwyl, and F.~Li.
\newblock Ferminets: Learning generative machines to generate efficient neural
  networks via generative synthesis.
\newblock 2018.

\bibitem{nas}
B.~Zoph, V.~Vasudevan, J.~Shlens, and Q.~V. Le.
\newblock Learning transferable architectures for scalable image recognition.
\newblock 2017.

\end{thebibliography}
}

\end{document}